\documentclass{llncs}
\usepackage{graphicx} 

\begin{document}
\title{Exploiting Reputation in Distributed Virtual Environments}
\author{Walter Quattrociocchi and Rosaria Conte}
\institute{University of Siena,Italy \\walter.quattrociocchi@unisi.it \\Istc-Cnr, Rome Italy\\rosaria.conte@istc.cnr.it}

\maketitle
\begin{abstract}
The cognitive research on reputation has shown several interesting properties that can improve both the quality of services and the security in distributed electronic environments.
In this paper, the impact of reputation on decision-making under scarcity of information will be shown.
First, a cognitive theory of reputation will be presented, then a selection of simulation experimental results from different studies will be discussed. Such results concern the benefits of reputation when agents need to find out good sellers in a virtual market-place under uncertainty and informational cheating.
\end{abstract}



\section{Introduction}

This paper is intended to collect and re-examine the properties of reputation, intended as a social artifact, in contexts where information is poor and uncertain. 
Since reputation has co-evolved with the human language \cite{dunbar98}, it should not lose its human nature even when the substrate on which interaction is implemented changes.

The dynamics of social systems depend upon interactions among individuals.  People can populate their informational domain, that can be defined as the amount of information available to a society members through gossip, by communicating opinions and acquired information to other people and, in turn, getting information from them. 

In this context, on the one hand, we have the individual source, based upon encounters, communications or observations that lead agents to form their opinions and social evaluations. 
On the other hand, the collective source generates information that is transmitted from one agent to another without the informers necessarily adopting it as truthful. 

It is a kind of ``meta-information'' which agents build up thanks to a special form of social intelligence, the capacity to form and manipulate the minds of other agents in their minds. 
This meta-information may be inaccurate, both because a great deal of information around the evaluation (who is the evaluator, when and which factors contributed to build this
evaluation) is lost and because the information is passed on without being checked out.
Hence, there is a need for agents to reduce, metabolize and compensate uncertainty. 

In the present work we show the positive effects produced by reputation in reducing the uncertainty when users are faced with decisions. The effects are elicited and characterized by means of multi-agent based simulations of a computational model (Repage) which implements a cognitive theory of reputation.

The paper is structured as follows: in the first section, a cognitive theory of reputation is introduced; in the second section, the cognitive computational model (Repage) of such theory and the simulation scenarios are discussed. Finally results of a series of simulations aimed at outlining the capabilities of reputation in decision making are provided.

\section{Reputation Theory}
In this section we introduce briefly a cognitive theory of reputation \cite{ContePaolucci} that is then simulated. More precisely, the reputation theory proposed in \cite{ContePaolucci} introduces a  fundamental distinction between \textit{image} and \textit{reputation} in social evaluations. 
Both are mental constructs, concerning other agents' (target) attitudes toward a given socially desirable behavior, that may be shared by a multitude of agents. 
Image is an evaluative belief and asserts that the target is ``good'' when it displays
certain behavior and ``bad'' in the opposite case.
Reputation differs from image because it is true when the evaluation conveyed is actually 
widespread but not necessarily accurate. 
Furthermore, reputation circulates within society through gossip (\cite{Coleman90}), that is a communication stream which allows, among other effects, cheaters to be found out or neutralised. 
To take place, reputation spreading needs four types of agents: evaluators, targets, gossipers and
beneficiaries. As an example, the sentence ‚ ``Silvio is considered an honest person'',
includes a set of evaluators (who adopt the evaluation), a single targets (Silvio), a set of
beneficiaries (the people for which it is important to know that Silvio is an honest
person) and the set of gossipers (who refer the evaluation as such, independent from its adoption).

Hence, to acknowledge the existence of a reputation does not imply to accept the evaluation itself. For instance, an agent $A$ might have a very good image of agent $B$ as a seller and at the same time recognize that there is a voice about agent $B$ as a bad seller.

Unlike ordinary communication and deception, reputation does not bind the speaker to
commit herself to the truth-value of the evaluation conveyed but only to the existence of
rumors about it. 
Hence, reputation implies neither personal commitment nor responsibility over the evaluation transmission 
and on its consequences. It represents the agents' opinions about a certain target's attitude with respect to a specific norm or standard or her possession of a given skill.
The reputation as described below has been adopted an studied in different contexts such as e-governance\cite{eRep}, and internet of services \cite{Balke08}, the reputation systems \cite{Josang2009}, and so forth.

\section*{Repage}
In this section we introduce the multi-agent based computational model developed from the theory introduced above. Many other reputation models have been proposed in the field of agent-based systems
such as in \cite{schillo99} \cite{regret}; for a comparison of these models
from the point of view of information representation, refer to \cite{Sabater2007}. 
However, in these models, image and reputation collapse on the same type of
social evaluation. Even when agents are allowed to acquire undirect information, this will
not be represented as distinct from and interacting with own opinion. As a consequence,
there is no way to distinguish known rumors from shared evaluations, a distinction
which is instead  central to our theoretical approach. 
Reputation co-evolved with human language and social organization as a multi-purpose social 
and cognitive artifact \cite{dunbar98}. It incentives cooperation and norm abiding, while discouraging defection and free-riding 
by (1) allowing retaliation against transgressors and (2) cooperating at the level of information exchange. 
The various aspects of this theory has been explored by means of multi-agent based simulation
with the Repage Framework \cite{Repage}. Repage provides evaluations on potential partners and 
is fed with information from others and outcomes from direct experience. 
To select good partners, agents need to form and update their own social
evaluations; hence, they must exchange evaluations with one another. 
If agents communicate only the believed image, the circulation of social knowledge 
would be bound to stop soon. But in order to preserve their autonomy, agents 
need to decide whether to share or not other's evaluations of a given target.
If agents transmit other's evaluations as if these evaluations
were their own, without the possibility of choosing if they
want to mix both types of evaluations or not, they would be
no more autonomous. Hence, they must

\begin{itemize}
 \item form both evaluations (image) and meta-evaluations
(reputation), keeping distinct the representation of own
and others' evaluations, before
\item deciding whether or not to integrate reputation with
their own image of a target.  
\end{itemize}

In Repage others can either transmit their own image of a given target,
which they hold to be true, or report on what they have heard about the target,
i.e. its reputation, whether they believe this to be true or
not. Of course, in the latter case, they will neither commit
to this information's truth value nor feel responsible for its
consequences. Consequently, agents are expected to transmit uncertain information, 
and a given positive or negative reputation may circulate over a population of agents even if
its content is not actually shared by the majority.

The main element of the Repage architecture is the memory (shown in Figure 1), 
which is composed by a set of predicates (\cite{quattrociocchi2007}). 

\begin{figure}[h!]
 \centering
 \includegraphics[width=60mm]{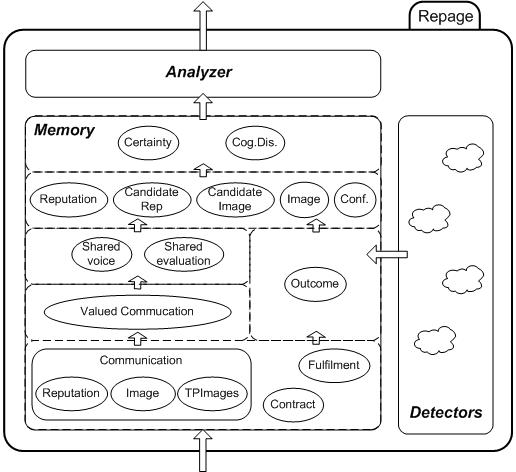}
 \label{fig:repageArch}
\caption{Regage Architecture}
\end{figure}

Predicates are objects containing a social evaluation, belonging to one
of the main types accepted by Repage (image, reputation,
shared voice, shared evaluation), or to one of the types used
for their calculation (valued information, evaluation related
from informers, and outcomes). Predicates contain a tuple
of five numbers representing the evaluation plus a strength
value that indicates the confidence the agent has on this evaluation. 
Predicates are conceptually organized in different
levels and connected by a network of dependencies, specifying 
which predicates contribute to the values of others. Each
predicate in the Repage memory has a set of antecedents and
a set of consequents. If an antecedent is created, removed,
or changes its value, the predicate is notified, recalculates its
value and notifies the change to its consequents. What is of interest
in the economy of the present discussion is that in order to
make use of Repage's complex network of dependencies we
need to aggregate social evaluations, by taking into account
what exactly is expressed by the social evaluations.

\section{The Virtual Market}
The experiments reported in this paper (\cite{quattrociocchi2007}, \cite{quattrociocchi2008},\cite{quattrociocchi2008b} \cite{quattrociocchi2010b}, \cite{quattrociocchi2009e} and \cite{quattrociocchi2010c}) were performed in a simulated marketplace in which agents 
can purchase goods by selecting a seller. 
This selection is facilitated by the possibility to acquire information from other agents about the quality of sellers or about the quality of other agents as informers. Answers are provided as shared (image) or reported upon (reputation) evaluations.

The market has been designed with the purpose of reproducing the simplest possible setting where information is both valuable and scarce. It includes only two kinds of agents, the buyers and the sellers. All agents perform actions in discrete time units (turns from now on). In a turn, a buyer performs one communication request and one purchase operation. In addition, the buyer answers all the information requests that it receives. Goods are characterized by a utility factor that we interpret as quality (but, given the level of abstraction used, could as well represent other utility factors as quantity, discount, timeliness) with values between 1 and 100. Sellers are characterized by a constant quality, drawn following a stationary probability distribution, and a fixed stock, that is decreased at every purchase; they are essentially reactive, their functional role in the simulation being limited to providing an abstract good of variable quality to the buyers. Sellers exit the simulation when the stock is exhausted and are substituted by a new seller with similar characteristics but with a new identity (and as such, unknown to the buyers). The disappearance of sellers makes information necessary; reliable communication allows to discover faster better sellers. This motivates the agents to participate in the information exchange. In a setting with permanent sellers (infinite stock), once all buyers have found a good seller, there is no reason to change and the experiment freezes. With finite stock, even after having found a good seller, buyers should be prepared to start a new search when the good seller's stock ends. At the same time, limited stock makes good sellers a scarce resource, and this constitutes a motivation for the agents not to distribute information. One of the interests of the model is in the balance between these two factors.

\section{Discovering Good Sellers}

If reputation is a means for finding out material cheaters, i.e. non-reciprocators in exchange of material goods, how can such a mechanisms be exploited to find out informational cheaters, i.e. deceitful informants? 
Solutions to this problem in the MAS field usually rely on learning strategies (see \cite{Yu03} and \cite{Fullam2006}), by means of which agents more or less gradually discover trustworthy informers and gain accurate information. 

Applied to detection of informational cheating, learning techniques are more efficient than in the case with material cheating, since facing deception in reputation is less costly than trading with a cheater. 
Here the number of sellers and buyers is fixed to 100 and 15, respectively. Goods are represented by a utility factor that we interpret as quality (but, given the level of abstraction used, could as well represent other utility factors as quantity, discount, timeliness) with values between 1 and 100. 
\begin{figure}[h]
 \centering
 \includegraphics[width=60mm]{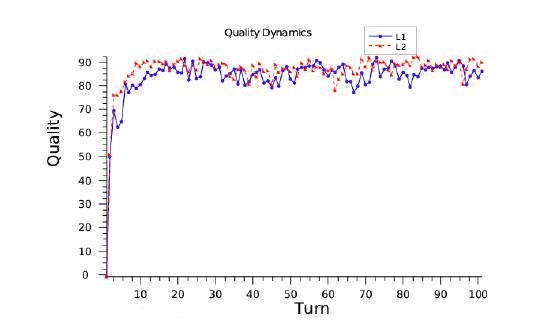}
 \caption{Average quality, against simulation ticks, for L1 (image only) and L2 (image and reputation). 
Both configurations achieve optimal quality, with faster L2 convergence}
 \label{fig:gsquality}
\end{figure}
\begin{figure}[h]
 \centering
 \includegraphics[width=60mm]{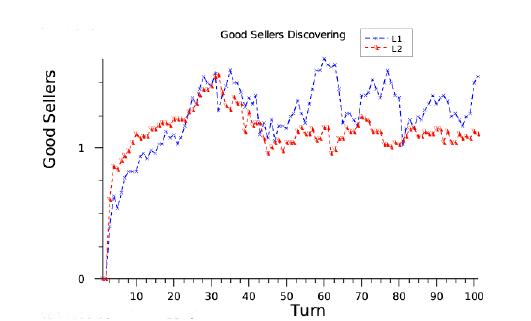}
 \caption{Good sellers discovered for each turn, L1 and L2. Note the larger spread and
periodical behavior in L1}
 \label{fig:gsDiscover}
\end{figure}

In Figure \ref{fig:gsquality} the quality of contracts in the simulated environment per simulation time is shown. In both scenarios (with only image and with image and reputation) the levels are equals.

On the contrary, the two curves present a different cyclic behavior: in L1 the the discovering of
good sellers come out by information exchange based only on image circulating
in the system.  The peaks of each wave in \ref{fig:gsDiscover} represents, according to the theory, that, once discovered a good seller, buyers start to buy from this one. 
Hence, the distribution of requests is not stable because of the finite number of stock for each
seller. 
In our view, the minimum value for each wave indicates a very slow process
of discovering when only image circulates in the system. This result suggests that reputation-based information has a stabilizing effect in a world of agents that communicate and learn to evaluate one another under inaccurate information.

\section{Reducing Uncertainty}
In this section the capacity of reputation to reduce uncertainty is discussed and contextualized.
The results are drawn by \cite{quattrociocchi2008} and \cite{quattrociocchi2008b}.
where the authors argue that not only a large amount of information
allows uncertainty reduction, but also that different sources
of information, in particular image and reputation, contribute to uncertainty reduction.
The hypotheses behind the simulation scenario can be summarized as follows:

\begin{itemize}
 \item in the domain of social and economic exchange, partner selection is fundamental to increase chances of cooperation and quality of products exchanged.
 \item  with image, we have partner selection with either retaliation, 
which partially neutralizes the good effect of partner selection, but with controlled information circulating, 
which reduces the quantity of information available for agents' use
\item  with reputation, we have partner selection with less
retaliation and more, although possibly uncertain, information circulating into the system
\item with reputation there is a larger quantity of information circulating in the system and we expect a decrease of uncertain evaluations, not only for the effect of the larger quantity of information, but also because meta-beliefs convert heterogeneous descriptions about a certain target into a synthetic evaluation.
\end{itemize}

\begin{figure}[h]
 \centering
 \includegraphics[width=80mm]{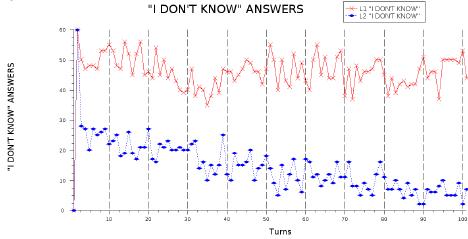}
 \caption{Trend of I-dont-know Answers in L1 and
L2}
 \label{fig:ur}
\end{figure}

When an agent in Repage has not yet formulated (neither by direct experience (image) nor by undirect information (reputation)) a complete evaluation about a given target, will provide an ``I-dont-know'' answer to question about the target.

In Figure \ref{fig:ur}, the trend of uncertainty, represented by answers of I-dont-know type, is shown for 100 simulation turns. The curves present a considerably different behavior: in L1, with only image circulating, after an initial faster growth values stabilize on high levels. This behavior is consistent with the reputation theory, the main expectation of which being that systems with only image circulating fail to decrease uncertainty. In L2, the trend of uncertain evaluations starts to decrease after some iterations until it reaches a lower value in correspondence with the spread of bad reputation.

\section{Avoiding Informational Cheating}

\begin{figure}[h!]
 \centering
 \includegraphics[width=80mm]{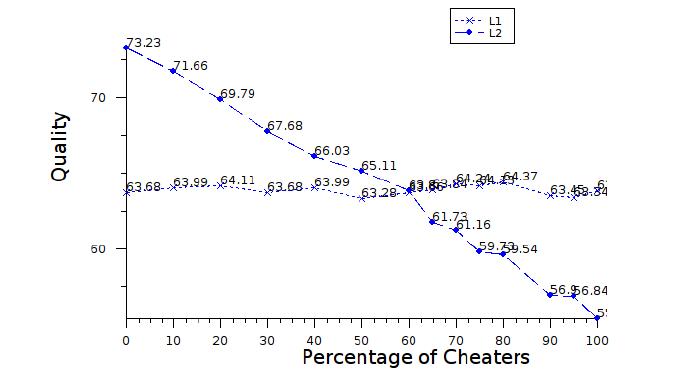}
 \caption{The curves represent quality in the stabilised regime for L1 and L2. Until cheaters
remain below the threshold of 60\%, reputation allows for quality to reach higher
values than happens in the complementary condition. The truth circulates faster
providing social evaluations which are coherent with the reality. Coherence
between information and reality, shared by a larger amount of agents in L2,
increases the trust and the market is more efficient. Over the threshold of 60
false information floods in, hence, social evaluations, circulating faster with
reputation, are often distorted with respect to reality.}
 \label{fig:cheating}
\end{figure}

The functioning of communication-based systems, as any socio-cognitive system, requires individuals to interact in order to acquire information enabling one to cope with uncertainty. These systems rely on the accuracy and on completeness of information. If agents
need to be correctly informed, then the quality of information becomes a fundamental factor of the global system dynamics. In order to obtain complete
information, agents need to include as many viewpoints as possibile in their perspectives. Different angles of observation ought to be presented as objectively as possible. What is the relation between
information credibility and opinion formation? What happens when the informational domains are not reliable and the trust level is poor? Can false information spread
throughout a society to such an extent that it becomes a prophecy? In the real world, each decision needs strategies to reduce the level of uncertainty in the process of beliefs' formation and revision with respect to the decision's consequences.
For instance, the theory of agenda-setting was introduced in 1972 by Maxwell McCombs and Donald Shaw in \cite{mcombs1972}. Their studies addressed the effect of media-delivered information on the presidential campaign of 1968 in Chapel Hill, North Carolina. Their theory states an existing correlation between information delivered by media with its consequent perceived importance. Despite the increasing diffusion of the Internet, which offers a large variety of different sources of information in a more interactive process of content fruition, traditional media still remain the major sources of information.
Information delivered by the media, in particular through the news, are perceived by the audience as reliable, as individuals are not in the position to check 
information quality. Consequently, the reliability of providers becomes a matter of trust.
The media are bound to exercise a profound cultural impact on society by affecting both the public perceptions of matters of general interest and the socalled agenda, the dynamics of public opinion. The media deeply affect the mental representations of facts and the revision of beliefs. Media-delivered information also impacts on the way in which facts are described. The successive elaboration of information, conveyed through gossip, inherits the vocabulary, as well as the underlying notions, values, and stereotypes transmitted by the media.
In the following picture (Figure \ref{fig:cheating}) the trends of the quality of contracts are shown. 
Here the experimental settings are designed in a way that the agent's objective is to reach 
the maximum quality value in a contract under informational cheating. 
The picture shows the average market quality in the stabilised regime in L1 and L2 for a percentage of cheaters varying from 0 to 100\%. We can observe the different performances of image and reputation at the increasing number of cheaters: reputation is more sensitive than image to the increase of cheating, but gives
a better average quality performance than image does, until the rate of informing cheaters
overcomes 60\%. For a comparative analysis, results are presented for the following critical values of the cheaters' number: at 0\% of cheaters, when only true information circulates; at 60\% when the positive effects of reputation disappears; and, finally, at 100\% when all of the agents are informational cheaters.

\section{Conclusions}
This paper was intended to collect and re-examine the properties of reputation as defined in \cite{ContePaolucci}, intended as a social artifact, in contexts where information is poor and uncertain. In particular, it was shown that reputation is a good means for finding out good sellers: by means of reputation uncertainty can be reduced and potential cheaters can be inhibited.
Since reputation has co-evolved with the human language, it should not lose its human nature even when the substrate on which interaction is implemented changes \cite{CFQS2010,ACFQS2010a}.
The theory and the model introduced in this paper should either guide the implementation of algorithms to compute reputation scores and be useful in analogous decisional settings.
In future works we will approach the definition of other social artifacts with the main aim to provide  detailed, cognitively-sound and non-reductionist computational models which are strictly related with social evaluations (e.g. opinions \cite{GiardiniQC2011a,GiardiniQC2011b} and their dynamics \cite{quattrociocchi2011a,quattrociocchi2010e}). 

\section{Acknowledgements}
A particular thanks to Geronimo Stilton and the Hypnotoad.

\bibliographystyle{plain}
\bibliography{biblio}

\end{document}